\documentclass{article}
\usepackage{spconf,amsmath,graphicx}
\usepackage{cite}
\usepackage{multirow}
\usepackage{multicol}
\usepackage{inconsolata}
\usepackage{capt-of}
\usepackage{tabularx}
\usepackage{booktabs}
\usepackage{scalerel}
\usepackage[export]{adjustbox}
\usepackage{pifont}
\usepackage{float}
\usepackage{amssymb}
\usepackage{color}
\usepackage[table]{xcolor}
\usepackage{caption}
\usepackage[numbers,sort&compress]{natbib}
\makeatletter
\renewcommand\subsection{\@startsection{subsection}{2}{\z@}%
                       {1.6ex}%
                       {0.6ex\@afterindentfalse}%
                       {\normalfont\large\bfseries}}
\makeatother

\makeatletter
\renewcommand\paragraph{\@startsection{paragraph}{4}{\z@}%
                       {0.45ex}%
                       {0pt}%
                       {\normalfont\normalsize\bfseries}}
\makeatother

\title{DialCLIP: Empowering CLIP as Multi-Modal Dialog Retriever}

\name{
        Zhichao~Yin$^{1}$\textsuperscript{\textdagger\textasteriskcentered}, Binyuan~Hui$^{3}$\textsuperscript{\textdagger}, Min~Yang$^{2}$\textsuperscript{\textdaggerdbl}, Fei~Huang$^{3}$, Yongbin~Li$^{3}$\textsuperscript{\textdaggerdbl}
}
\address{
        $^{1}$University of Science and Technology of China\\
        $^{2}$Shenzhen Institute of Advanced Technology, Chinese Academy of Sciences\\
        $^{3}$DAMO Academy, Alibaba Group\\
        \{zc.yin,min.yang\}@siat.ac.cn,
        \{binyuan.hby,shuide.lyb\}@alibaba-inc.com
}

\begin{document}
\maketitle

\renewcommand{\thefootnote}{\fnsymbol{footnote}}
\footnotetext[1]{Work done while the author was interning at Alibaba DAMO Academy.}
\footnotetext[2]{Equal contribution.}
\footnotetext[3]{Corresponding authors.}
\renewcommand{\thefootnote}{\arabic{footnote}}

\begin{abstract}
Recently, substantial advancements in pre-trained vision-language models have greatly enhanced the capabilities of multi-modal dialog systems. These models have demonstrated significant improvements by fine-tuning on downstream tasks. However, the existing pre-trained models primarily focus on effectively capturing the alignment between vision and language modalities, often ignoring the intricate nature of dialog context. In this paper, we propose a parameter-efficient prompt-tuning method named DialCLIP for multi-modal dialog retrieval. Specifically, our approach introduces a multi-modal context prompt generator to learn context features which are subsequently distilled into prompts within the pre-trained vision-language model CLIP. 
Besides, we introduce domain prompt to mitigate the disc repancy from the downstream dialog data.
To facilitate various types of retrieval, we also design multiple experts to learn mappings from CLIP outputs to multi-modal representation space, with each expert being responsible to one specific retrieval type. Extensive experiments show that DialCLIP achieves state-of-the-art performance on two widely recognized benchmark datasets (i.e., PhotoChat and MMDialog) by tuning a mere 0.04\% of the total parameters. These results highlight the efficacy and efficiency of our proposed approach, underscoring its potential to advance the field of multi-modal dialog retrieval.
\end{abstract}

\section{Introduction}
Developing a highly effective multi-modal dialog system that is capable of providing informative and contextually appropriate responses has long been a goal of artificial intelligence research. Existing multi-modal dialog systems can be broadly classified into two categories: generation-based approaches\cite{shuster2020multi,MMGen} and retrieval-based approaches\cite{imagechat,zang2021photochat,mmdialog}. Generation-based systems, exemplified by large language models such as GPT-4\cite{OpenAI2023GPT4TR}, generate responses in an end-to-end manner. While these models have shown impressive results, retrieval-based approaches remain indispensable in certain scenarios. The retrieval-based approach enables dialog systems to produce controlled responses, particularly in security- or privacy-sensitive situations.

\begin{figure}[t!]
    \centering
    \setlength{\belowcaptionskip}{-3.5ex}
    \setlength{\abovecaptionskip}{0ex}
    \includegraphics[width=\columnwidth]{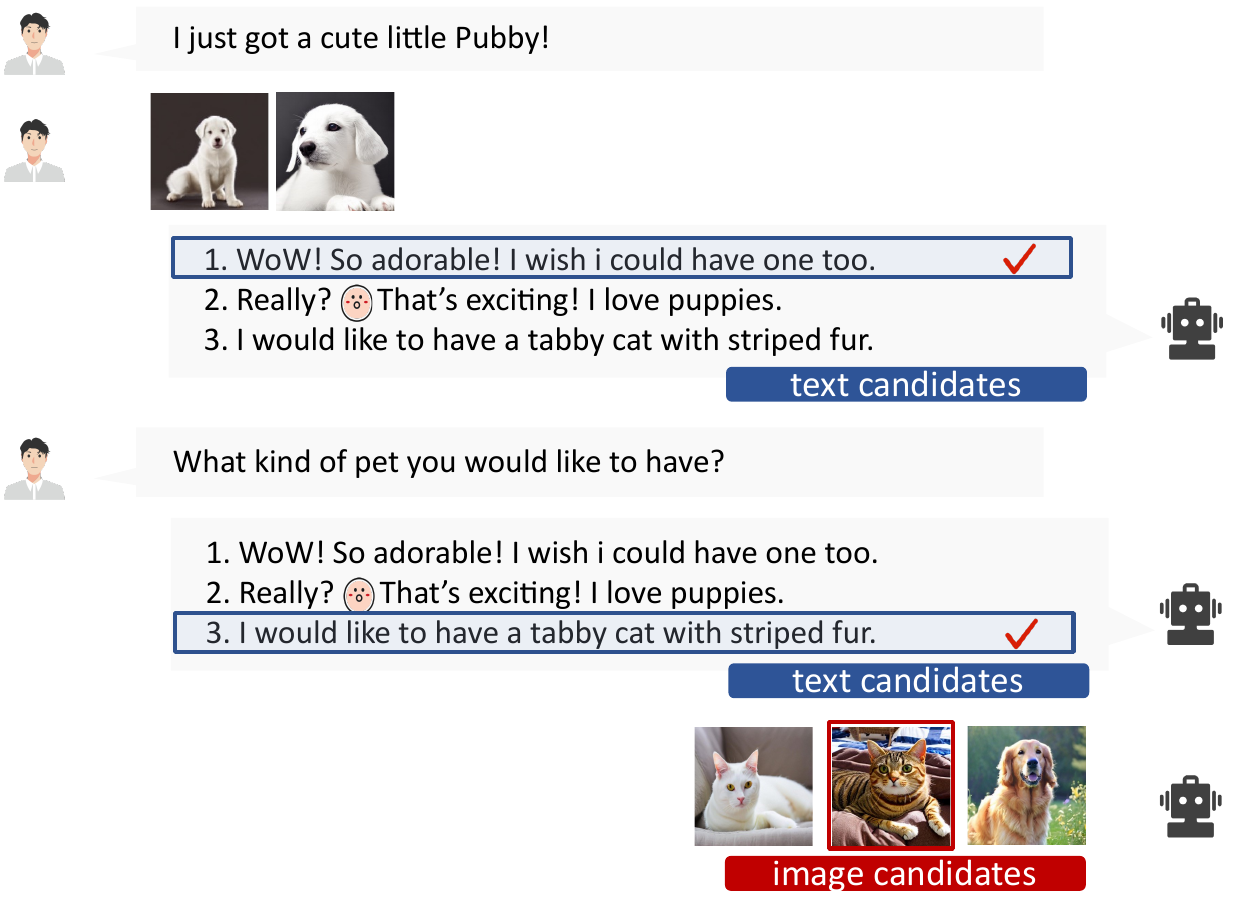}
    \caption{Example of retrieve-based multi-modal dialog system.}
    \label{mm-instance}
\end{figure}

The recent advancements in pre-trained vision-language models~\cite{radford2021learning,alayrac2022flamingo,ofa,wang2023image} have significantly enhanced the performance of multi-modal dialog systems by further fine-tuning these models on specific downstream tasks. However, existing pre-trained multi-modal models primarily focus on modeling the alignment between vision and language, often neglecting the modeling of dialog context. As illustrated in Figure \ref{mm-instance}, in real-life applications, dialog context can be multi-turn and multi-modal content, involving both user and system inputs. Effectively capturing and modeling the multi-modal dialog history becomes crucial in generating semantically and contextually appropriate responses.

One potential solution is to leverage the power of pre-trained vision-language model CLIP~\cite{radford2021learning}. However, directly applying CLIP to multi-modal dialogs presents three primary dilemmas.
\textbf{First}, since the pre-training of CLIP relies on single-round pairwise data without explicitly considering the rich multi-modal dialog context. It requires a concise and efficient way to inject multimodal contexts.
\textbf{Second}, CLIP is trained using non-dialog data, which limits its usefulness in dialog-based tasks.
\textbf{Third}, multi-modal dialog involves different types of response selection, not only text-to-image. The projection scheme CLIP uses is not fit for this scenario.


In this paper, we propose DialCLIP, a model that leverages efficient prompt tuning~\cite{brown2020language,scalingpeft,liu2021gpt,jin2022instance,latetuning} to empower CLIP as a multi-modal dialog retriever.
\textbf{First}, we design a multi-modal context encoder that aligns language and image representations within their respective contexts. Outputs of the context encoder are then distilled into prompts to enhance multi-modal contextual awareness of CLIP. 
\textbf{Second}, we introduce domain prompts to adapt CLIP to dialog retrieval tasks.
\textbf{Third}, we propose different projection experts to tackle different retrieval types with each expert handling one specific type.
Extensive experiments demonstrate DialCLIP's effectiveness.
Specifically, the IR@1 metric on MMDialog is significantly improved by 12.8, and the Sum metric on PhotoChat is improved by 17.8, achieving state-of-the-art results. 

\begin{figure*}[t!]
    \centering
    \setlength{\belowcaptionskip}{-2ex}
    \setlength{\abovecaptionskip}{0ex}
    \includegraphics[width=\linewidth]{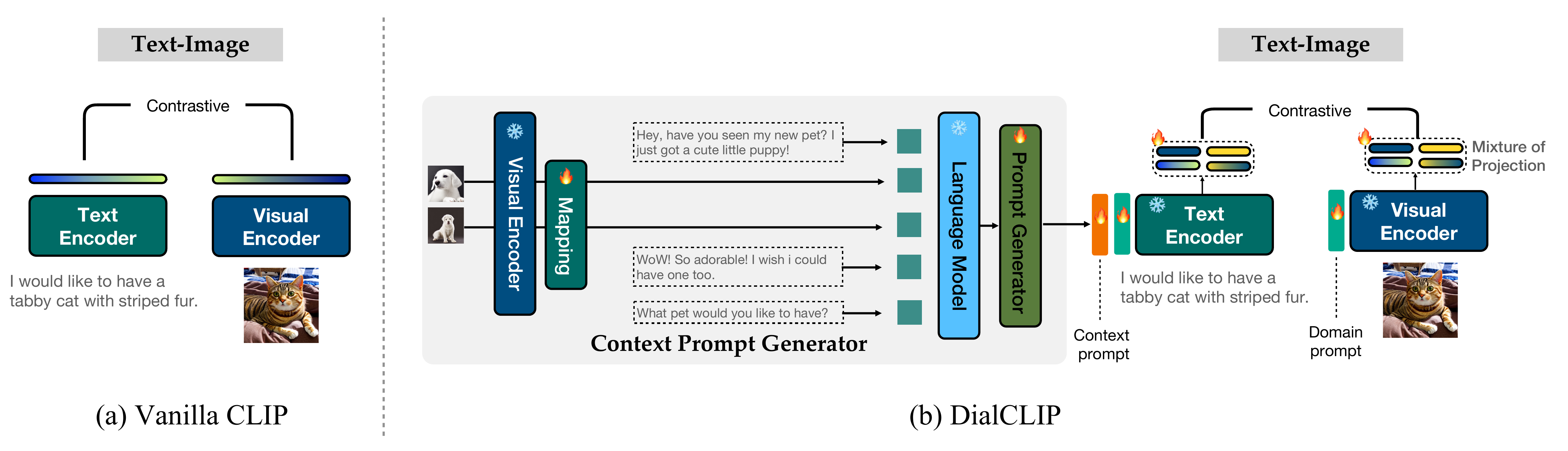}
    \caption{The architecture of our multi-modal response retrieval model (called DialCLIP), which consists of context prompt generator (CPG),  CLIP encoders, and mixture of projection (MoP).} 
    \label{arch}
\end{figure*}

\vspace{-2ex}
\section{Method}
\vspace{-0.5ex}
\subsection{Task Definition}
Given dialog history $H=\{U_0^m, U_1^m, U_2^m,...,U_{n-1}^m, U_n^m\}$, the goal is to select the most relevant response $R^m$. Different from the traditional textual dialogs, both $U_i^m$ and $R^m$ could be textual content or visual image under the multi-modal dialog scenario. $m \in \{t, v\}$ indicates the modal type of elements where $t$ represents textual utterances while $v$ represents visual images. In this work, we take the utterances of previous $n$-1 rounds as the multi-modal dialog context $C$ and the query of the $n$-th round as the current input $I$.

\subsection{Model Architecture}
As illustrated in Figure \ref{arch}, our multi-modal response retrieval model consists of a context prompt generator,  CLIP encoders and mixture of projection. The context prompt generator takes multi-modal context as input, and generate context prompts using prompt generator. The features encoded by CLIP encoders are then mapped to the common space using the according projection expert.

\paragraph*{Context Prompt Generator}~
To convert dialog context into prompts, we build context prompt generator (CPG), which consists of a multi-modal context encoder and a prompt generator. Inspired by BLIP2~\cite{li2022blip}, we develop our multi-modal context encoder by bridging the modality gap between pre-trained language encoder and visual encoder with a linear mapping network. After encoding the dialog context, we use prompt generator to distill the output features into prompt tokens $P_c$. Prompt Generator is composed of a pooling layer to control the length of prompts and a feed-forward neural network to convert output features into CLIP-compatible context prompts:
\begin{align}
    & \hat{P}_c(\mathbf{h}) = {\rm ReLU}(W_1({\rm Pooling}(\mathbf{h})) + b_1)) \\
    & P_c(\mathbf{h}) = W_2(\hat{P}_c(\mathbf{h})) + b_2 
\end{align}
where $W_1 \in \mathbb{R}^{m \times d}$ and $W_2 \in \mathbb{R}^{d \times m}$ are learnable parameters, and $\mathbf{h} \in \mathbb{R}^{d \times n}$ represents encoded context features. $n$ is the length of the original input. 

\paragraph*{Domain Prompt}~
CLIP is pretrained using a large corpus of images paired with their corresponding natural language descriptions. However, there are significant differences between the image-text pairs used in CLIP and the dialogue data. To help CLIP capture knowledge of dialog, we introduce a special set of parameters, called domain prompt. Domain prompt is designed to improve the generalization performance of CLIP on dialog retrieval tasks. We implement domain prompt using deep prompt tuning~\cite{li2021prefix}, where fixed-length learnable parameters are added to each layer of the CLIP encoders.
\paragraph*{Mixture of Projection}~
CLIP maps the representations from each encoder to the multi-modal representation space by using single projection scheme. However, in the multi-modal dialog retrieval scenario, single projection is properly not sufficient since there could be multiple retrieval types according to the modalities of the last utterance and the response. To accommodate different retrieval types, we adopt multiple experts for projection, named Mixture of Projection (MoP). In the remaining parts, we denote the prompted CLIP encoders with MoP as $\tilde{E}$.
\begin{align}
\tilde{E}(H) = {\rm MoP}({\rm CLIP_{t/v}}(P_c;P_d;I)) \\
\tilde{E}(R^m) = {\rm MoP}({\rm CLIP_{t/v}}(P_d;R^m))
\end{align}
where $P_c$ and $P_d$ denote context prompts and domain prompts respectively. 

\subsection{Model Training}
In our method, the multi-modal dialog context $C$ is first passed through the context encoder $E_c$ to obtain the context prompts $P_c$ using the prompt generator. The context prompts $P_c$ along with the domain prompts $P_d$ are then inserted into the current input and encoded into a $D$-dimensional vector $\mathbf{x}$ using the enhanced CLIP encoder. Similarly, the response $R^m$ is also encoded into a $D$-dimensional vector $\mathbf{y}$. Once we have the representations of the dialog history and the responses, we can compute their relevance scores between $\mathbf{x}$ and $\mathbf{y}$ using the dot-product similarity.
\begin{align}
& s(\mathbf{x},\mathbf{y}) = \mathbf{x} \cdot \mathbf{y}
\end{align}

During training, our objective is to minimize the distance between the utterance history and the ground-truth response in the feature space, while simultaneously maximizing the distance between the utterance history and irrelevant responses. This encourages the model to accurately capture the essence of the conversation and differentiate between relevant and irrelevant responses. We define the overall contrastive loss function as follows:
\begin{multline}
\mathcal{L}(\mathbf{x}_i, \mathbf{y}_i^+, \{\mathbf{y}_j^-\}_{j=1}^N) = \\
- \log \frac{ e^{s(\mathbf{x}_i, \mathbf{y}_i^+)} }{e^{s(\mathbf{x}_i, \mathbf{y}_i^+)} + {\sum_{j=1}^{N} e^{s(\mathbf{x}_i, \mathbf{y}_j^-)}} }
\end{multline}
where $\mathbf{x}_i$ indicates the representation of the $i$-th training sample, $\mathbf{y}^{+}_i$ denotes the representation of the positive response of $\mathbf{x}_i$, $\mathbf{y}^{-}_j$ denotes the representation of the $j$-th negative response of $\mathbf{x}_i$, and $N$ is the number of training samples. 

\begin{table*}[t!]
    \centering
    \setlength{\abovecaptionskip}{0.5ex}
    \setlength{\belowcaptionskip}{-0.5ex}
    \resizebox{0.84\textwidth}{!}{%
    \begin{tabular}{lcccccccccc}
    \toprule
    \multirow{2}{*}{Models} & \multicolumn{6}{c}{MMDialog} & \multicolumn{4}{c}{PhotoChat} \\
    \cmidrule(lr){2-7} \cmidrule(lr){8-11}
    & TR@1 & TR@5 & TR@10 & IR@1 & IR@5 & IR@10 & R@1 & R@5 & R@10 & Sum \\
    \midrule
    DE++\cite{mmdialog} & 22.2 & 39.4 & 47.6 & 29.8 & 48.2 & 57.6 & 8.3  & 25.8  & 36.2  & 70.3\\
    CLIP\cite{radford2021learning} & 24.3 & 43.1 & 48.2 & 28.7 & 50.4 & 59.9 & 12.6  & 33.2  & 42.6  & 89.6\\
    VLMo\cite{bao2022vlmo}& 29.0 & 42.7 & 52.3 & 29.3 & 51.6 & 63.5 & 13.8 & 30.0 & 39.4 & 83.2\\
    PaCE\cite{pace}& 29.1 & 45.9 & 56.2 & 34.6 & 59.2 & 71.3 & 15.2 & 36.7 & 49.6 & 101.5\\
    \cmidrule(lr){1-11}
    \textbf{DialCLIP} & \textbf{34.9} & \textbf{56.0} & \textbf{64.5} & \textbf{47.4} & \textbf{72.0} & \textbf{80.3} & \textbf{19.5} & \textbf{44.0} & \textbf{55.8} & \textbf{119.3}\\
    \bottomrule
    \end{tabular}%
    }
    \caption{
    Performance comparison between DialCLIP and baseline models. 
    }
    \label{tab:model-performance}
\end{table*}

\begin{table*}[t!]
\setlength{\abovecaptionskip}{0.5ex}
\setlength{\belowcaptionskip}{-2ex}
\centering
    \begin{minipage}{0.31\textwidth}
    \resizebox{0.98\textwidth}{!}{
    \begin{tabular}{ccccc}
    \toprule
    Length & R@1 & R@5 & R@10 & Sum \\
    \midrule
    16 & 6.0 & 18.2 & 28.4 & 52.6 \\
    32 & 8.5 & 24.7 & 36.6 & 69.8 \\
    64 & 11.5 & 29.0 & 40.6 & 81.1 \\
    \rowcolor{gray!30}
    96 & 12.2 & \textbf{32.3} & 42.4 & \textbf{86.9}\\
    128 & \textbf{12.9} & 30.6 & \textbf{43.2} & 86.7 \\
    \bottomrule
    \end{tabular}
    }
    \caption{Performance on PhotoChat with different context prompt lengths.}
    \label{tab:ctx-len}
    \end{minipage}
    \hspace{3px}
    \begin{minipage}{0.31\textwidth}
    \resizebox{0.98\linewidth}{!}{
    \begin{tabular}{ccccc}
    \toprule
    Length & R@1 & R@5 & R@10 & Sum \\
    \midrule
    2 & 12.0 & \textbf{33.7} & 44.8 & 90.5 \\
    \rowcolor{gray!30}
    4 & \textbf{13.6} & 32.5 & \textbf{45.9} & \textbf{92.0} \\
    8 & 13.1 & 33.2 & 45.3 & 91.6 \\
    16 & 9.5 & 29.0 & 42.0 & 80.5 \\
    64 & 8.7 & 28.6 & 39.8 & 77.1 \\
    \bottomrule
    \end{tabular}
    }
    \caption{Performance on PhotoChat when varying domain prompt lengths.}
    \label{tab:dom-len}
    \end{minipage}
    \hspace{3px}
    \begin{minipage}{0.31\textwidth}
    \resizebox{0.98\linewidth}{!}{
    \begin{tabular}{ccccc}
    \toprule
    Layer & R@1 & R@5 & R@10 & Sum \\
    \midrule
    \rowcolor{gray!30}
    2 & \textbf{12.4} & \textbf{31.3} & \textbf{44.5} & \textbf{88.2} \\
    4 & 11.7 & 30.3 & 41.3 & 83.3 \\
    6 & 11.1 & 30.5 & 42.7 & 84.3 \\
    8 & 11.4 & 30.5 & 41.3 & 83.2 \\
    10 & 10.2 & 29.2 & 40.6 & 80.0 \\
    \bottomrule
    \end{tabular}
    }
    \caption{Performance on PhotoChat when varying prompt layers.}
    \label{tab:layer}
    \end{minipage}
\end{table*}

\section{Experiments}
\subsection{Experimental Setup}
\paragraph*{Datasets}~
We assess DialCLIP on two popular benchmarks: PhotoChat\cite{zang2021photochat} and MMDialog\cite{mmdialog}. PhotoChat is the first dataset which proposes photo sharing task. MMDialog is currently the largest multimodal dialogue dataset. We use Recall as the automatic evaluation and report R@1, R@5, and R@10 scores for both PhotoChat and MMDialog. Since MMDialog requires replying with multimodal response, we report separate scores for text retrieval and image retrieval, denoted as TR and IR respectively.

\paragraph*{Baseline Methods}~
We adopt 4 baseline models for comparison. (i) VLMo\cite{bao2022vlmo} is a unified VLP model that leverages the multi-modal multi-expert (MoME) architecture; (ii) PaCE\cite{pace} stands as the pioneering pre-trained model specifically tailored for multi-modal dialogue tasks; (iii) DE++\cite{mmdialog} capitalizes on the CLIP framework to encode both textual utterances and visual images; and (iv) CLIP\cite{radford2021learning} learns representations of vision and language jointly by training on millions of image-text pairs. 

\paragraph*{Implementation details}~
In the context prompt generator, we employ the flan-t5-xl\cite{chung2022scaling} model as the language encoder and CLIP ViT-B/32 as the vision encoder. In our main experiments, we maintain a fixed length of 4 for the domain prompts, while setting the length of the context prompts to 96 for PhotoChat and 64 for MMDialog. The context prompts are inserted into the 2nd layer for PhotoChat  and the 6th layer for MMDialog, instead of the typical word embedding layer. During the training phase, we utilize the AdamW~\cite{loshchilov2017decoupled} optimizer and implement a linear learning rate scheduler, starting with a base learning rate of 5e-5. To initiate the training process, we sample 100k multi-modal pairs from the MMDialog and initially pre-train the matching heads and domain prompts for 16k steps, employing a batch size of 256. Subsequently, we continue training the DialCLIP model with a batch size of 160 for PhotoChat and 12 for MMDialog.

\subsection{Main Results}
Table \ref{tab:model-performance} presents the performance comparison, which showcases the outstanding performance of our method, surpassing previous approaches on both the PhotoChat and MMDialog datasets. Notably, despite having significantly fewer learnable parameters, DialCLIP exhibits a substantial performance improvement over fully fine-tuned methods. Specifically, DialCLIP achieves a notable improvement of 17.8 points (from 101.5 to 119.3) on the PhotoChat and a significant boost of 12.8 points (from 34.6 to 47.4) on the MMDialog.

\subsection{Analysis}
\paragraph*{Impact of Context Prompt Length}~
The impact of varying the length of context prompts on the model's performance intrigues our curiosity. We conduct a series of experiments using five different context prompt lengths: 16, 32, 64, 96 and 128 (while domain prompt length and prompt layer are fixed to 0). As table \ref{tab:ctx-len} shows, longer context prompts generally get better performance. However, we also observe a performance drop when the length is increased to 128. Although longer context prompts can better preserve detailed semantic information, it doesn't mean longer is always better. In fact, longer context prompts can also introduce unexpected noise, leading to a decrease in performance.

\paragraph*{Impact of Domain Prompt Length}~
We also investigated the impact of different domain prompt lengths: 2, 4, 8, 16 and 64 (while context prompt length and prompt layer are fixed to 96 and 0 respectively). The results are reported in Table \ref{tab:dom-len}. Interestingly, we observe a significant performance drop when the length of the domain prompt exceeds the threshold of 4. Furthermore, setting the length to 64 results in a dramatic decline in the model's performance. The reason behind this phenomenon is that long domain prompts can cause overfitting on the training data. Moreover, long domain prompts may also dilute the impact of context prompts due to the attention mechanism. 

\paragraph*{Effect of Prompt Layers}~
We also conduct a comprehensive analysis to investigate the impact of inserting context prompts at different intermediate layers, aiming to identify the optimal position for inserting context prompts (while context prompt length and domain prompt length are fixed to 96 and 0 respectively). Table \ref{tab:layer} shows our experiment results. Surprisingly, inserting context prompts into the embedding layer does not yield the best performance. In fact, setting the prompt layer to the 2nd layer results in an increased performance, reaching 88.2 on PhotoChat. These results highlight the criticality of carefully selecting the position of the prompt layer to achieve optimal performance.

\paragraph*{Effect of Initialization}~
The initialization has a great impact on the performance. We compare three initialization approaches and report their results in Table \ref{tab:initialization}: (i) all parameters randomly initialized (called Random); (ii) pretrain the domain prompts and MoP first (called Pretrain); and (iii) initiate the domain prompts using task-related words (e.g. "response"), called Task-related. To our surprise, using relevant words for initialization yields slightly inferior results compared to random initialization. However, we observe that pretraining has a profound positive impact on the performance. In the absence of pretraining, the performance on the PhotoChat dataset experiences a significant decrease of 26.6 points. 

\begin{table}[h]
    \centering
    \setlength{\abovecaptionskip}{0.5ex}
    \setlength{\belowcaptionskip}{-3ex}
    \resizebox{0.76\linewidth}{!}{
    \begin{tabular}{lcccc}
    \toprule
    Method & R@1 & R@5 & R@10 & Sum \\
    \midrule
    Random & 13.8 & 32.6 & 46.3 & 92.7 \\
    \textbf{Pretrain} & \textbf{19.5} & \textbf{44.0} & \textbf{55.8} & \textbf{119.3} \\
    Task-related & 13.6 & 32.5 & 44.5 & 90.6 \\
    \bottomrule
    \end{tabular}
    }
    \caption{
    The performance of DialCLIP on PhotoChat by using different initialization methods.
    }
    \label{tab:initialization}
\end{table}

\paragraph*{Ablation Study}~
Table~\ref{tab:ablation} shows ablation results on the MMDialog benchmark. As it illustrates, DialCLIP performs worst when without CPG, which indicates CPG can provide useful contextual information within generated prompts.

\begin{table}[h]
    \centering
    \setlength{\abovecaptionskip}{0ex}
    \setlength{\belowcaptionskip}{-4.5ex}
   \resizebox{0.88\linewidth}{!}{
    \begin{tabular}{lcccccc}
    \toprule
    Method & TR@1 & TR@5 & TR@10 & IR@1 & IR@5 & IR@10 \\
    \midrule
    \textbf{DialCLIP} & \textbf{34.9} & \textbf{56.0} & \textbf{64.5} & \textbf{47.4} & \textbf{72.0} & \textbf{80.3}\\
    - CPG & 16.3 & 32.7 & 41.2 & 21.0 & 34.3 & 48.2 \\
    - Domain & 32.6 & 53.1 & 61.7 & 45.8 & 69.2 & 77.3 \\
    - MoP & 31.2 & 52.8 & 62.3 & 44.9 & 68.3 & 76.6 \\
    \bottomrule
    \end{tabular}
    }
    \caption{
    Results of ablation experiments on the MMDialog.
    }
    \label{tab:ablation}
\end{table}

\section{Conclusion}
\vspace{-0.5ex}
This paper investigates the efficacy of prompt tuning in the context of the multi-modal response retrieval task. We extend CLIP to a multi-modal dialog retrieval model using combined prompts. In addition, we perform extensive analysis to provide valuable insights into prompt design.

\vspace{-3ex}
\section*{Acknowledgments}
\vspace{-1.8ex}
Min Yang was supported by National Key Research and Development Program of China (2022YFF0902100), National Natural Science Foundation of China (62376262), Shenzhen Science and Technology Innovation Program (KQTD20190929172835662), Shenzhen Basic Research Foundation (JCYJ20210324115614039 and JCYJ202001091-13441941). This work was supported by Alibaba Group through Alibaba Innovative Research Program.
\bibliographystyle{IEEEbib}
\bibliography{main.bib}

\end{document}